\documentclass[runningheads]{llncs}
\usepackage{graphicx}
\usepackage{comment}
\usepackage{amsmath,amssymb} 
\usepackage{color}
\usepackage{subcaption}
\usepackage{multirow}
\usepackage{xcolor}
\usepackage{wrapfig}
\usepackage{ifthen}
\usepackage[ruled,vlined]{algorithm2e}
\usepackage{etoolbox}
\BeforeBeginEnvironment{wrapfigure}{\setlength{\intextsep}{0pt}}

\newboolean{combined}
\setboolean{combined}{false}
\newboolean{acknowledgement}
\setboolean{acknowledgement}{true}

\DeclareMathOperator*{\argmax}{argmax} 

\begin{document}

\pagestyle{headings}
\mainmatter
\def\ECCVSubNumber{33}  

\title{Improved Robustness to Open Set Inputs via Tempered Mixup} %

\titlerunning{Tempered Mixup} 
\authorrunning{R. Roady et al.} 

\author{Ryne Roady$^{1}$ \quad Tyler L. Hayes$^{1}$ \quad Christopher Kanan$^{1,2,3}$}

\institute{
Rochester Institute of Technology, Rochester, NY 14623 USA \and 
Paige, New York, NY 10044 USA \and 
Cornell Tech, New York, NY 10044 USA\\
  {\tt\small \{rpr3697,tlh6792,kanan\}@rit.edu}}

\maketitle

\begin{abstract}
Supervised classification methods often assume that evaluation data is drawn from the same distribution as training data and that all classes are present for training.  However, real-world classifiers must handle inputs that are far from the training distribution including samples from unknown classes. Open set robustness refers to the ability to properly label samples from previously unseen categories as novel and avoid high-confidence, incorrect predictions. Existing approaches have focused on either novel inference methods, unique training architectures, or supplementing the training data with additional background samples. Here, we propose a simple regularization technique easily applied to existing convolutional neural network architectures that improves open set robustness without a background dataset.  Our method achieves state-of-the-art results on open set classification baselines and easily scales to large-scale open set classification problems.
\end{abstract}

\section{Introduction}
Modern supervised classification methods often assume train and test data are drawn from the same distribution and all classes in the test set are present for training.  However, deployed models will undoubtedly be exposed to out-of-distribution inputs that do not resemble training samples and these models are expected to robustly handle these novel samples.  Performance in this `open-world' setting is often hidden by current computer vision benchmarks in which the train and test sets have the same classes and the data is sampled from the same underlying sources. One solution to this problem is to develop open set classifiers which have the ability to identify novel inputs that do not belong to any training classes so that they are not assigned an incorrect label~\cite{Scheirer2013Towards}. This capability is especially important for the development of safety-critical systems (e.g., medical applications, self-driving cars) and lifelong learning agents that automatically learn during deployment~\cite{parisi2019continual}.

\begin{figure}[t]
    \centering
    \includegraphics[width=\textwidth]{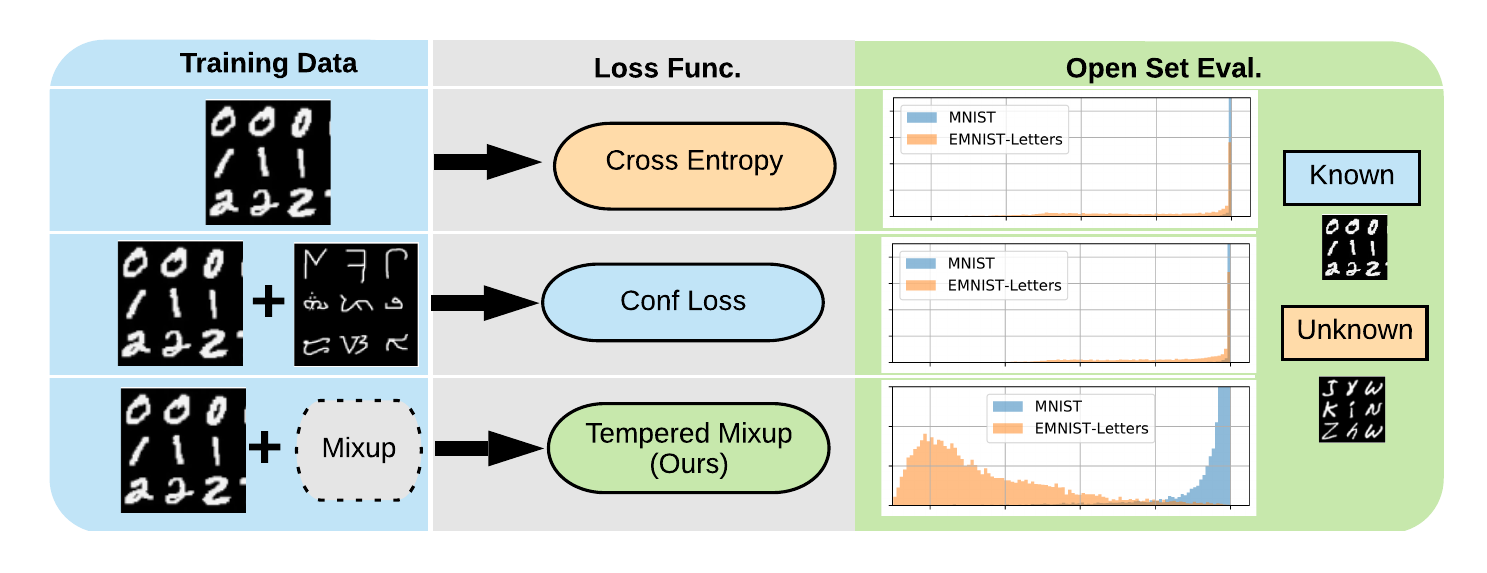}
    \caption{ \textsc{Tempered Mixup}. Traditional methods for building model robustness to unknown classes involves training with a background set, which is problematic for large-scale datasets.  Tempered Mixup provides state-of-the-art robustness to novel inputs without the requirement of a representative background set.  The model confidence of a LeNet++ model trained to classify MNIST digits is shown for both known and unknown classes here represented by the Extended-MNIST-Letters dataset. Confidence is calculated using only the softmax of the model output and does not rely on a computationally expensive open set inference method to separate known from unknown classes.}
    \label{fig:intro_fig}
\end{figure}

There are two major paradigms for enabling open set classification in existing convolutional neural network (CNN) architectures. The first paradigm replaces the standard closed-set prediction scheme with a new inference mechanism~\cite{bendale2015towards,liang2018enhancing,lee2018simple,shu2017doc}. The second paradigm is to regularize the classification model during training to enable it to better separate known classes from potential unknowns~\cite{lee2017training,dhamija2018reducing}. For the latter approach, the most effective methods involve training with a large set of background images to penalize an overconfident prediction when a sample from a novel or unknown class is encountered.  This approach has been shown to excel at open set classification, but existing results have been limited to small-scale datasets. In part, this is because it is increasingly difficult to construct an effective background dataset that does not semantically overlap with a large-scale training dataset.

In this paper, we overcome this limitation by proposing a new training approach that penalizes overconfident predictions on samples outside of the known training classes without access to a set of background/unknown inputs. Instead our method hallucinates invalid images using a novel form of the Mixup data augmentation technique~\cite{zhang2017Mixup} that we combine 
with a unique auxiliary loss function. \textbf{This paper makes the following contributions:}
\begin{enumerate}
    \item We propose a novel end-to-end training algorithm for regularizing existing CNN architectures for open set classification that does not require the use of an explicit background dataset. 
    \item We propose a new loss function specifically designed to train a model to be less confident towards samples from unknown classes.
    \item We show that our method exceeds methods that use explicit background datasets on standard small-scale benchmark settings for open set classification (e.g., MNIST and CIFAR).
    \item We demonstrate that our method can easily scale to open set classification on large-scale datasets and produce comparable results to background set regularization without having to build an additional dataset for training.
\end{enumerate}

\section{Background}

\subsection{Open Set Classification}
The goal of open set classification is to explicitly label points that are `far' from known training classes as an unknown class instead of arbitrarily assigning one of the known categories. Open set classification methods discriminate among $K$ known categories seen during training and reject samples that do not belong to the known categories. Formally, given a training set $D_{train} = \left\{ {\left( {{{X}}_1 ,y_1 } \right),\left( {{{X}}_2 ,y_2 } \right), \ldots ,\left( {{{X}}_n ,y_n } \right)} \right\}$, where ${X}_i$ is the $i$-th training input tensor and $y_i \in C_{train} = \left\{1, 2, \ldots, K  \right\} $ is its corresponding class label, the goal is to learn a classifier $F\left(X\right) = (f_1,...,f_k)$, that correctly identifies the label of a known class and separates known from unknown examples: 
 \begin{equation}
 \label{eq:bounded-classification-formulation}
     \hat{y} = \left\{\begin{array}{ll}\argmax_{k} F(X) & \mbox{if } {S}({X}) \geq \delta \\ K+1 & \mbox{if }{S}({X}) < \delta \end{array} \right.
 \end{equation}
where ${S} \left({X} \right)$ is an acceptance score function that determines whether the input belongs to the training data distribution, $\delta$ is a user-defined threshold, and $K+1$ indicates `unknown class.'  For proper open set testing, the evaluation set contains samples from both the set of classes seen during training and additional unseen classes, i.e., $D_{test} = \left\{ {\left( {{{X}}_1 ,y_1 } \right),\left( {{{X}}_2 ,y_2 } \right), \ldots ,\left( {{{X}}_n ,y_n } \right)} \right\}$, where $y_i \in  \left( C_{train}  \bigcup C_{unk} \right)$ and $C_{unk}$ contains classes that are not observed during training.

\subsection{Inference Methods for Open Set Classification}
Inference methods incorporate open set classification abilities into a pre-trained CNN by creating a unique acceptance score function and threshold that rejects novel inputs~\cite{Scheirer2013Towards,bendale2015towards,bendale2016towards}. Current state-of-the-art inference methods often rely on features from multiple layers of the CNN~\cite{lee2018simple,abdelzad2019detecting} and many methods use multiple forward and backward passes through the CNN to improve performance~\cite{liang2018enhancing,lee2018simple}.  These approaches significantly increase computational and memory requirements during inference~\cite{roady2019out}, which may be sub-optimal for deployed models.  We instead focus on building better open set classification performance through model regularization so that, during inference time, a much simpler and computationally efficient method such as confidence thresholding can be used to detect unknown classes. 

\subsection{Confidence Loss Training for Open Set Classification}
One approach for improving open set robustness is training with a confidence loss penalty~\cite{lee2017training,dhamija2018reducing}, which improves detection of unknown classes by penalizing overconfident predictions on samples that are outside the training classes.  As previously  observed~\cite{dhamija2018reducing,roady2019out}, inputs from unknown classes tend to be centered around the origin and have a smaller magnitude than samples from known classes in the deep feature space of a well-trained CNN .

Given an effective background dataset that is representative of unknowns expected to be seen during deployment, confidence loss training~\cite{dhamija2018reducing} collapses novel samples toward the origin of the deep feature space, resulting in lower confidence model outputs. The regularization penalty is a simple addition to standard cross-entropy loss during training, i.e.,
\begin{equation}
\mathcal{L}_{conf} = \left\{\begin{array}{ll}-\mathop{\mathrm{log}} S_k(F(X)) & \mbox{if } X \in D_{train}  \\ -\frac{1}{K} \sum_{k=1}^K \log S_k(F(X)) & \mbox{if } X \in D_{bkg}  \end{array} \right.
\label{eqn:conf_loss}
\end{equation}
where the first term is a standard cross-entropy loss for known classes ($D_{train}$) and the second loss term forces the model to push samples from a representative background class ($D_{bkg}$) toward a uniform posterior distribution.

\begin{wrapfigure}[29]{r}{0.6\textwidth} 
\centering
\includegraphics[width=0.6\textwidth]{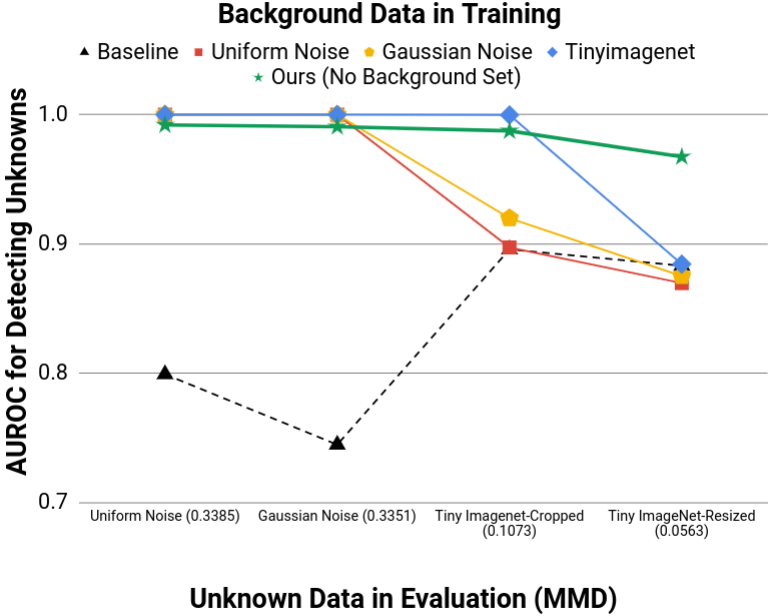}
\caption{\textsc{Novelty Detection vs Background Class Selection} The performance of confidence loss for improving open set performance is dependent on the selected background class.  As the background class becomes more dissimilar from the in-distribution data, it becomes less useful for improving model performance on difficult open set classification problems.  Our method does not require a background class for training and thus is robust across a wide-range of different input types.}
\label{fig:bkg_selection}

\end{wrapfigure}

Multiple approaches have been developed for producing open set images including using alternative datasets which are distinct from the training set~\cite{dhamija2018reducing,perera2019learning} and even using generative methods to produce images outside of the manifold defined by the training set~\cite{lee2017training,neal2018open}.  Naturally, the performance of the open set classifier trained with open set images is tied to the ``representativeness" of the background dataset and its similarity to the known training set.  The difficulty for even small-scale classification is that the effectiveness of certain background sets cannot be known a priori and seemingly representative background datasets can sometimes fail to produce better open set robustness.  We demonstrate this phenomenon using the CIFAR-10 dataset in Fig.~\ref{fig:bkg_selection}.  To overcome this limitation, we designed our method around drawing novel samples for confidence loss using data augmentation instead of relying on an explicit background set.

\subsection{Mixup Training}
Mixup is a regularization approach that combines two separate images from the training set into a single example by forming an elementwise convex combination of the two input samples~\cite{zhang2017Mixup}. Mixup can improve model accuracy~\cite{zhang2017Mixup}, model calibration~\cite{thulasidasan2019Mixup}, and model robustness to certain types of image corruptions~\cite{chun2019empirical}. However, Mixup has not been shown to be beneficial for open set classification, and in~\cite{chun2019empirical} Mixup resulted in a $50\%$ reduction in detecting unknown classes versus baseline cross-entropy training.  

Mixup is based on the principle of Vicinal Risk Minimization (VRM)~\cite{chapelle2001vicinal} where a classification model is trained not only on the direct samples in the training set, but also in the \textit{vicinity} of each training sample to better sample the training distribution.  In Mixup, these vicinal training samples $\left(\tilde{x} , \tilde{y} \right)$ are generated as a simple convex combination of two randomly selected input samples, $x_i$ and $x_j$:
\begin{equation} \label{eq:Mixup}
    \begin{split}
        \tilde{x} = \lambda x_i+(1-\lambda)x_j \\
        \tilde{y} = \lambda y_i+(1-\lambda)y_j
    \end{split}
\end{equation}
where $y_i$ and $y_j$ are the associated targets for the selected input samples.  The linear interpolation factor $\lambda \in [0,1]$ is drawn from a symmetric $Beta(\alpha,\alpha)$ distribution where the shape of the distribution is determined by the hyper-parameter, $\alpha$ which trades off training with mostly unmixed examples versus training with averaged inputs and labels.  By training with standard cross-entropy loss on these vicinal examples, the model learns to vary the strength of its output between class manifolds.  The effect of this training is a substantial improvement in model calibration and accuracy on large-scale image classification tasks~\cite{zhang2018Mixup,thulasidasan2019Mixup,chun2019empirical}. Our approach uses a variant of Mixup to overcome the need for an explicit background set with confidence loss training. 


\section{Mixup and Open Set Classification}
\subsection{Re-balancing Class Targets}
As originally proposed, Mixup is a form of model regularization that trains on linear combinations of inputs and targets, and encourages a model to learn smoother decision boundaries between class manifolds.  While smoother decision boundaries promote better generalization, they also benefit open set classification. This is because smoother boundaries reduce the likelihood of a model producing a confident but wrong prediction with an input that does not lie on the class manifolds learned from the training set.  Mixup essentially turns a single label classification problem into a multi-label problem by creating additional samples through the linear mixture of inputs and feature space embeddings. The question then becomes: is a linear combination of the targets appropriate for a linear combination of features in training a model to produce accurate uncertainty estimates in the space between class manifolds? 

\begin{figure}[t]
    \centering
    \begin{subfigure}{0.48\textwidth}
        \includegraphics[width=0.99\textwidth]{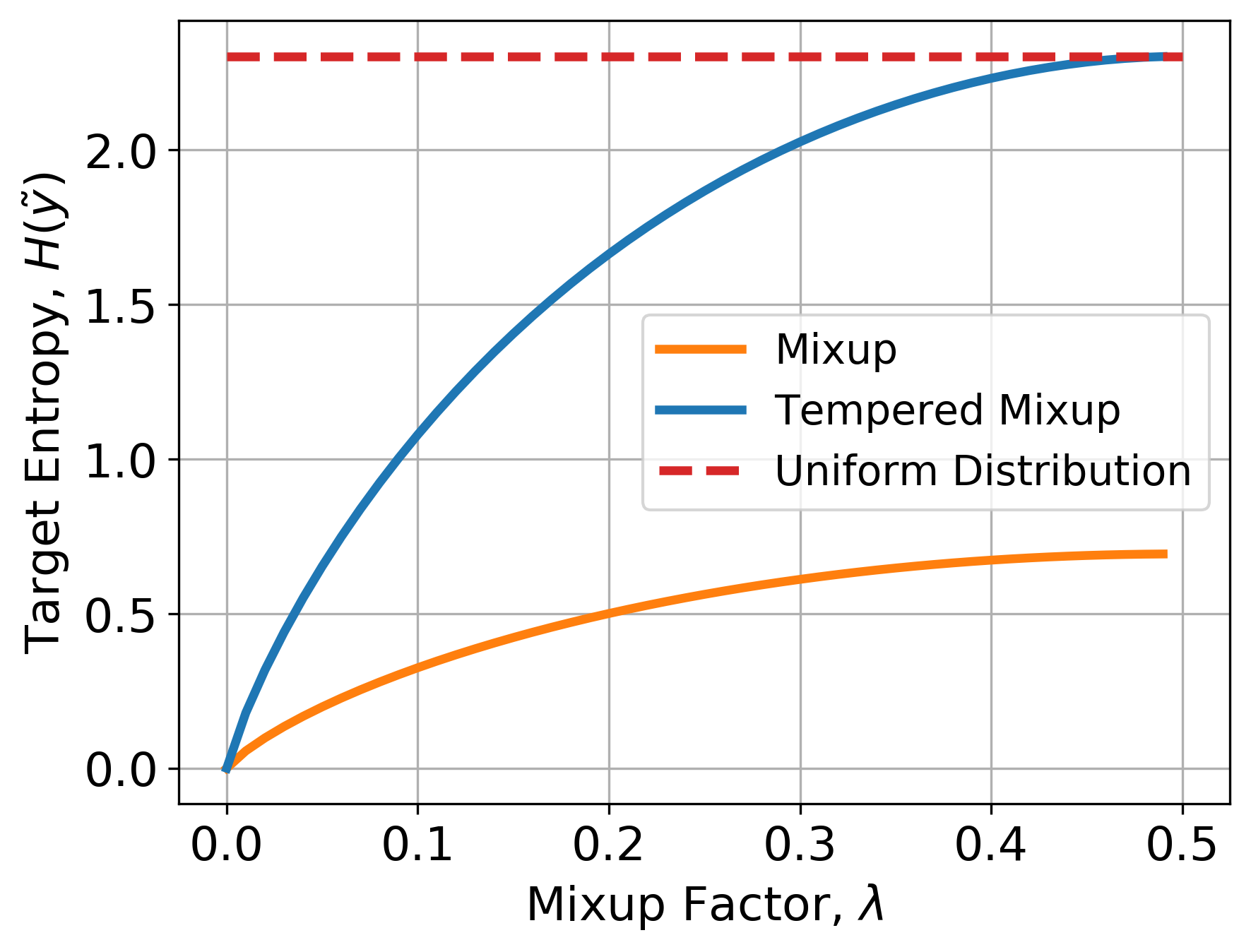}
        \caption{10-class dataset \\ (e.g. MNIST)}
        \end{subfigure} \hfill
    \begin{subfigure}{0.48\textwidth}
        \includegraphics[width=0.99\textwidth]{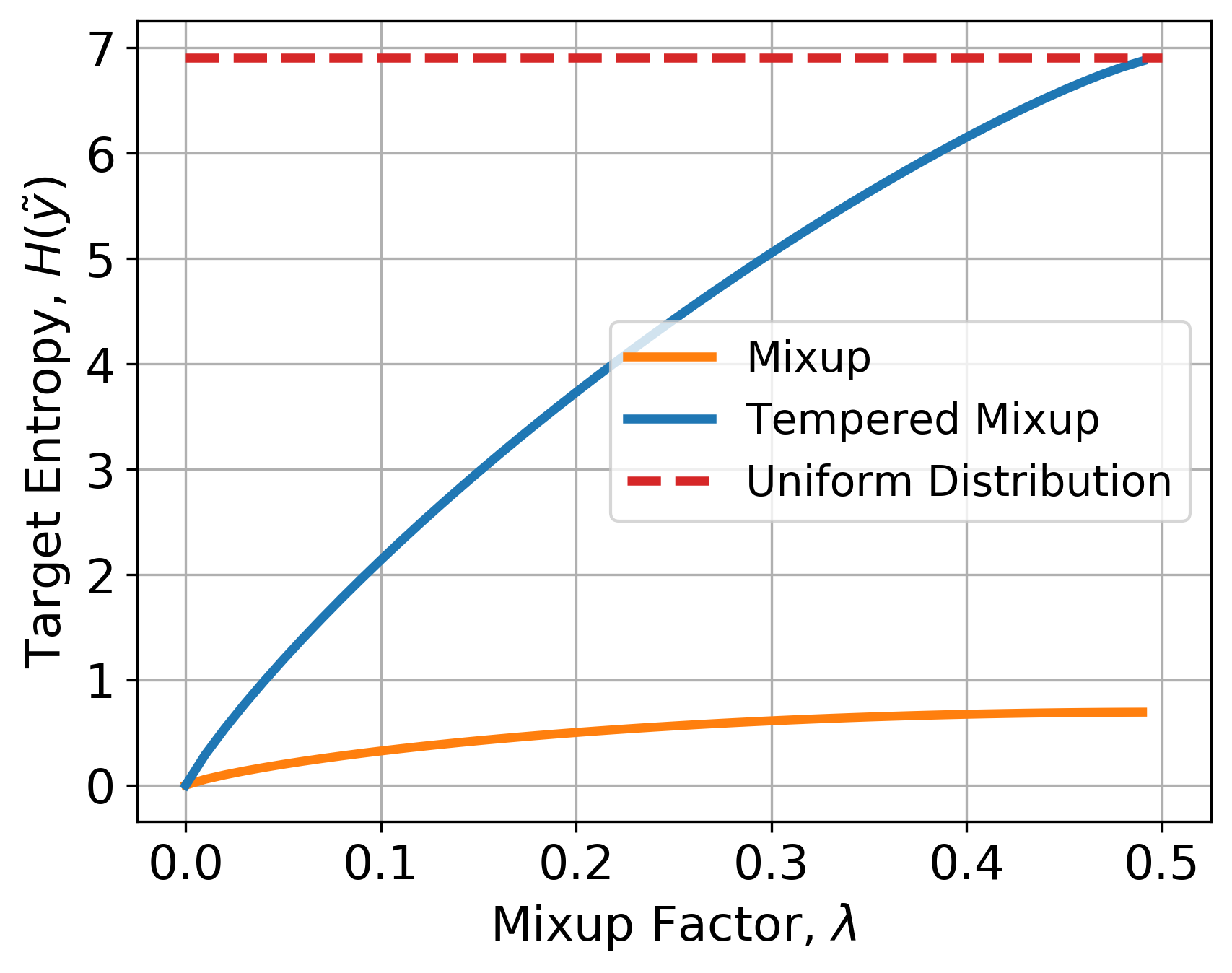}
        \caption{1000-class dataset \\ (e.g. ImageNet)}
        \end{subfigure}
    \caption[Mixup target rebalancing]{\textsc{Target Rebalancing}: Ideally, for unknown inputs a classifier's output probabilities will approach a uniform distribution. Using Mixup, samples that are more mixed should have higher entropy, but this does not occur to the extent necessary in the original formulation. Instead, Tempered Mixup uses a novel formulation that ensures the target entropy for mixed examples approaches that of a uniform distribution. This figure demonstrates this in both the 10 class and 500 class setting.
    }
    \label{fig:entropy}
\end{figure}

We answer this question by looking at how the target entropy of a model trained via cross-entropy loss changes as a function of the mixing factor $\lambda$.  As shown in Fig.~\ref{fig:entropy}, using a linear combination of targets, Eq.~\ref{eq:Mixup}, to mix the labels does not capture the increase in uncertainty that we desire for examples that are off of the class manifold, e.g., the highly mixed examples.  Instead we can re-balance the target labels with an additional label smoothing term modulated by the interpolation factor, $\lambda$, with an adjusted target mixing scheme as follows:
\begin{equation} \label{eq:target_mix}
        \tilde{y}_{tm} = |2\lambda -1| \tilde{y} + \frac{1-|2\lambda -1|}{K}\ ,
\end{equation}
where $\tilde{y}$ is the normal linear mixing from Eq. \ref{eq:Mixup} and $K$ is the number of known target classes.  Using this novel re-balancing approach, we can temper the model confidence for highly mixed samples such that they approach a uniform distribution prediction.  As shown in Fig.~\ref{fig:entropy}, this approach assigns a much higher target entropy to highly mixed up samples, including when they are mixed from the same class.  This tempering effect is magnified as the number of known classes increases.

\subsection{Tempered Mixup}
Tempered Mixup is an open set classification training method that overcomes the need for a background training set by using a modified form of Mixup with a novel variant of confidence loss regularization. Using Mixup enables the creation of \textit{off-manifold} samples based on the training input distribution, and it enables control over how similar the simulated outliers are to the known classes. This allows the CNN to learn features that are robust to open set classes through a number of different specialized inference methods, including baseline confidence thresholding.

Instead of training with standard cross-entropy (softmax) loss with labels drawn from the convex combination of two randomly selected images as prescribed in the standard Mixup algorithm, Tempered Mixup uses the same mixed up input but a modified version of the auxiliary confidence loss function to regularize how the model maps these between-class inputs in deep feature space. To do this, we apply the Mixup coefficient drawn per sample from a symmetric Beta distribution to a modified confidence loss equation. This allows us to simultaneously minimize the loss for misclassifying samples from the known classes and map unknown samples that are far from the known classes to the origin. The Tempered Mixup loss is given by:
\begin{equation} \label{eq:tempered_Mixup_loss}
\mathcal{L}_{TempMix} = -|2\lambda -1|\sum_{k=1}^K\tilde{y}_k \log \sigma_{S}(F(\tilde{X}))_k - \zeta \frac{1-|2\lambda -1|}{K} \sum_{k=1}^K \log \sigma_{S}(F(\tilde{X}))_k,
\end{equation}
where $\sigma_S$ is the softmax function applied to the vector $F(\tilde{X})$, $\lambda$ is the sampled mixing interpolation factor, $\tilde{y}$ denotes the linearly mixed targets, $K$ is the number of known classes, and $\zeta$ weights the amount of confidence loss applied to highly mixed up samples. Tempered Mixup is a straight-forward extension of traditional mini-batch stochastic gradient descent with cross-entropy loss training for deep neural network models.

\subsection{Visualizations of Deep CNN Feature Space to Unknowns} \label{sec:LeNet}

To visually illustrate the benefit of Tempered Mixup in separating known and unknown samples, we trained a simple CNN model (LeNet++ architecture~\cite{lecun1998gradient}) to classify MNIST digits as known classes and Extended MNIST Letters as unknown classes (overlaid as black points).  The CNN architecture has a bottle-necked two dimensional feature space to allow the visualization of the resulting embeddings.  

\begin{figure}[t]
    \centering
    \begin{subfigure}[b]{0.25\textwidth}
        \centering
        \includegraphics[width=\textwidth]{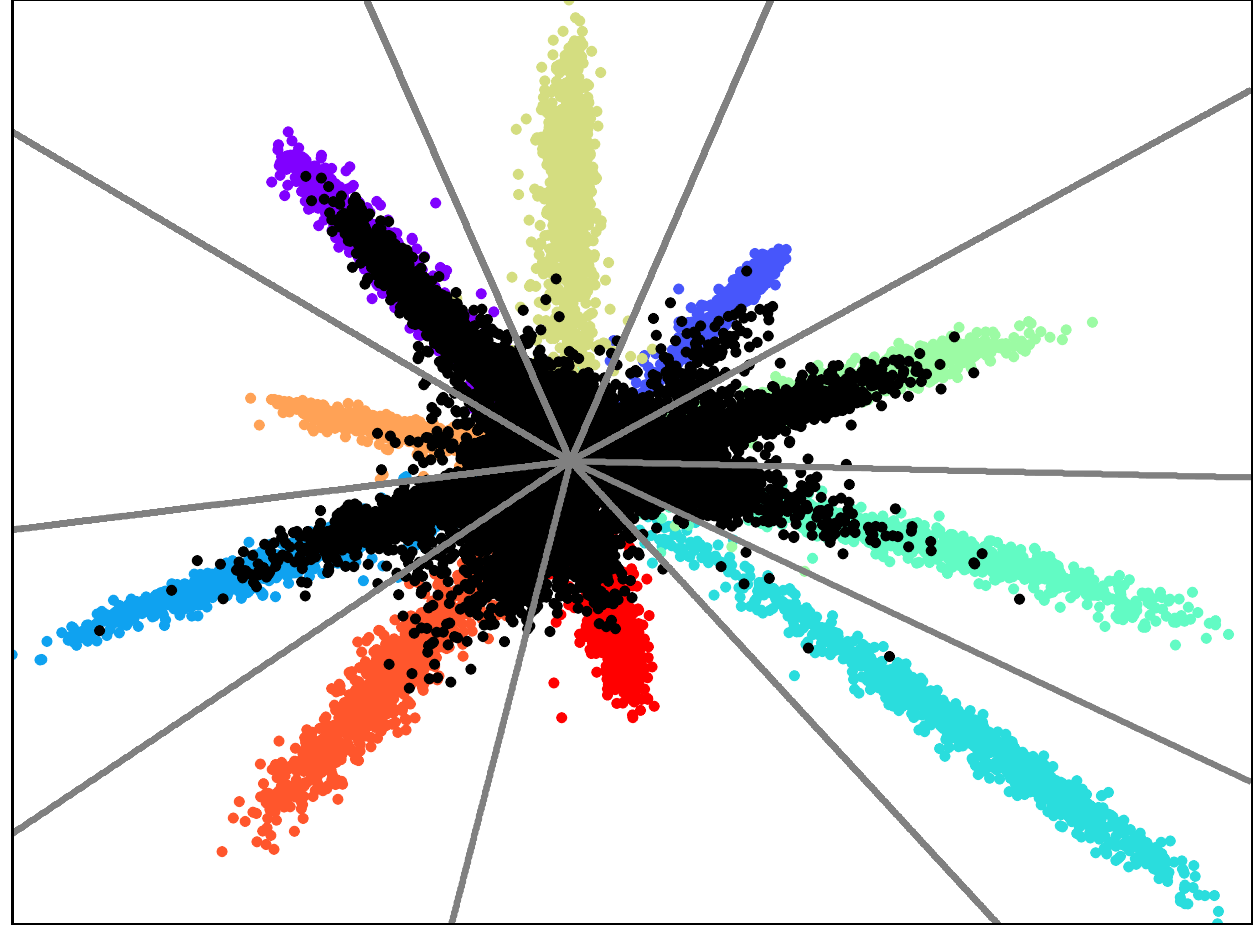}
        \caption{Cross-Entropy}
        \label{fig:lenet_baseline}
    \end{subfigure}
    \begin{subfigure}[b]{0.25\textwidth}
        \centering
        \includegraphics[width=\textwidth]{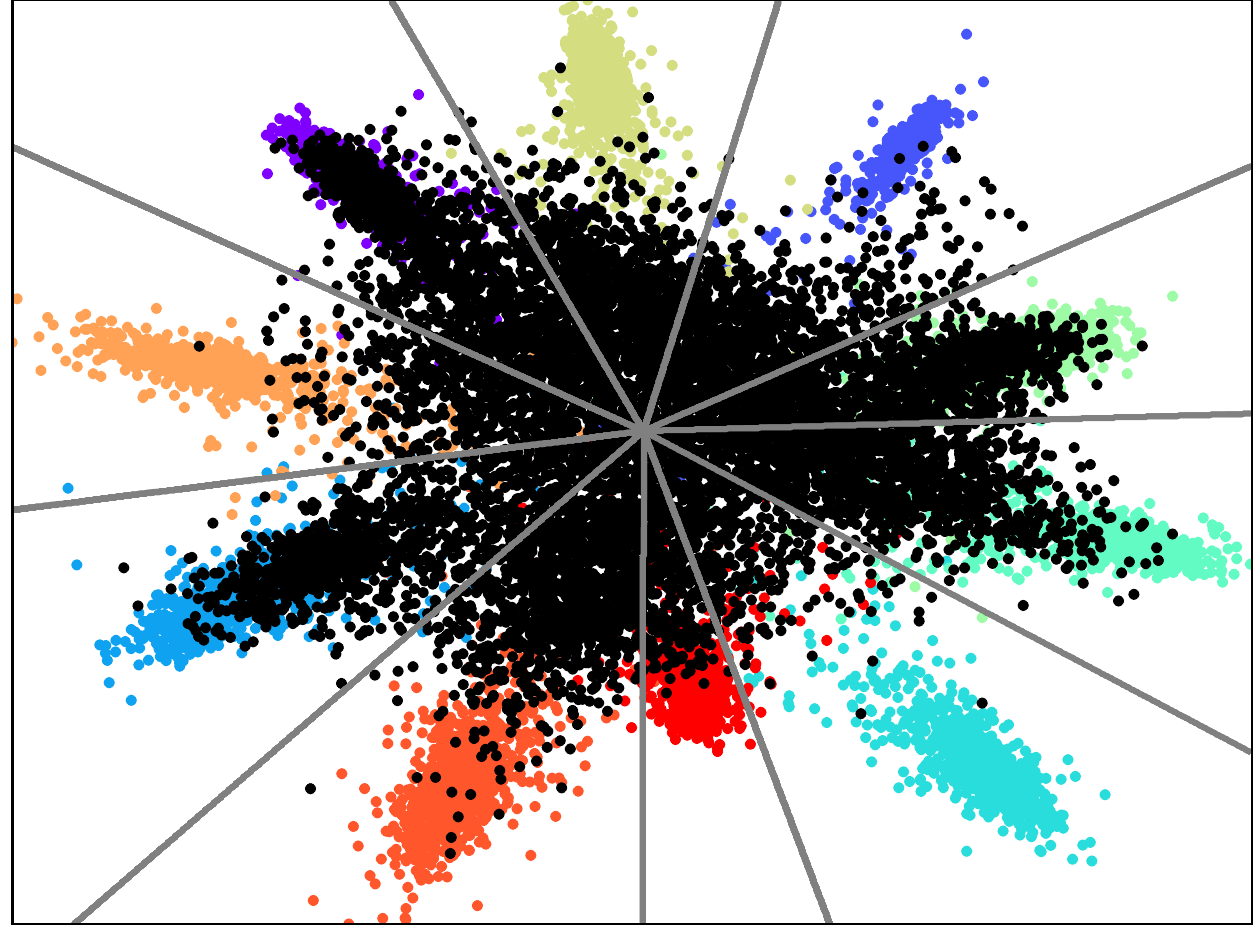}
        \caption{Label Smoothing}
        \label{fig:lenet_label_smooth}
    \end{subfigure}
    \begin{subfigure}[b]{0.25\textwidth}
        \centering
        \includegraphics[width=\textwidth]{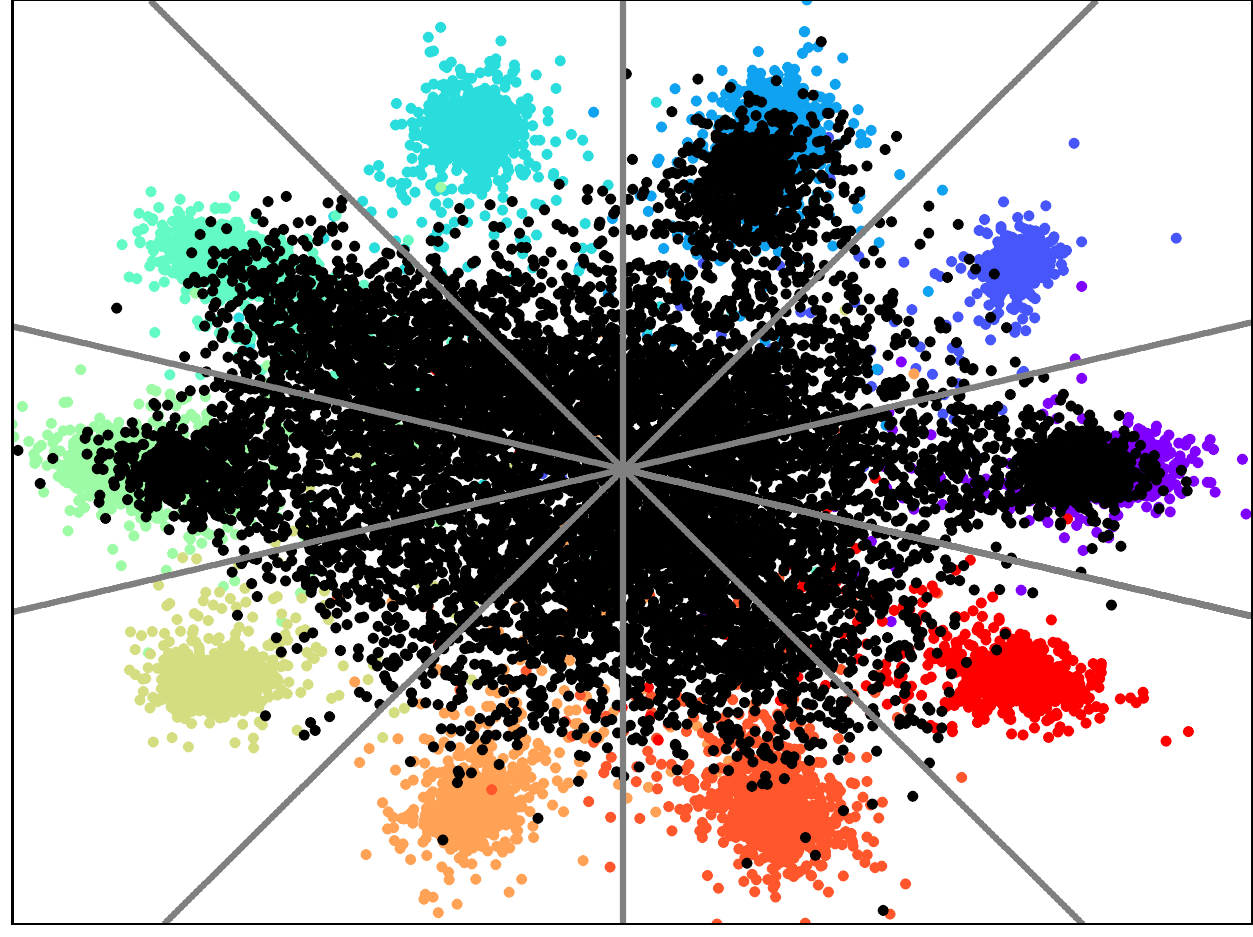}
        \caption{One-vs-Rest}
        \label{fig:lenet_one_vs_rest}
    \end{subfigure}
    \begin{subfigure}[b]{0.25\textwidth}
        \centering
        \includegraphics[width=\textwidth]{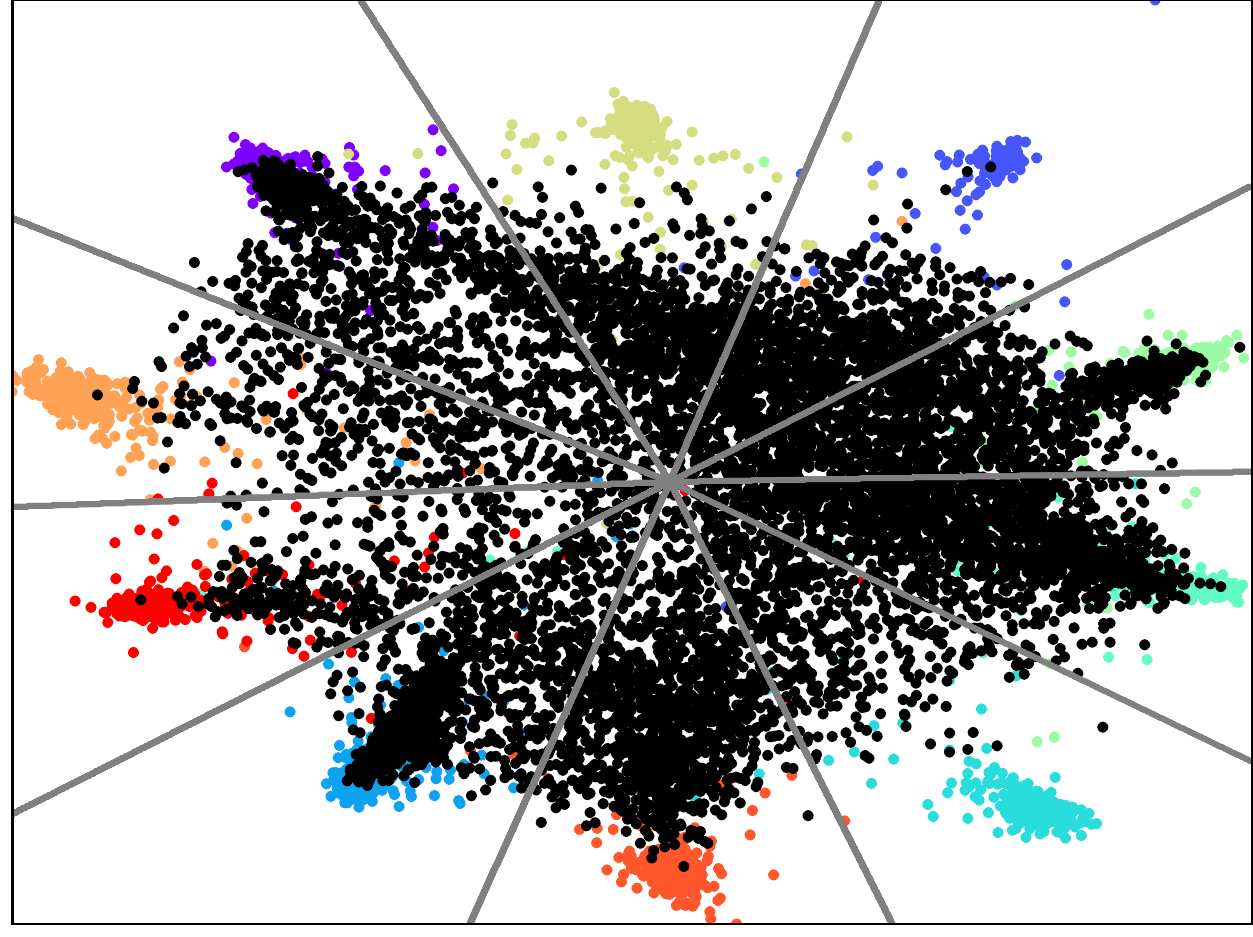}
        \caption{Center Loss}
        \label{fig:lenet_centerloss}
    \end{subfigure}
        \begin{subfigure}[b]{0.25\textwidth}
        \centering
        \includegraphics[width=\textwidth]{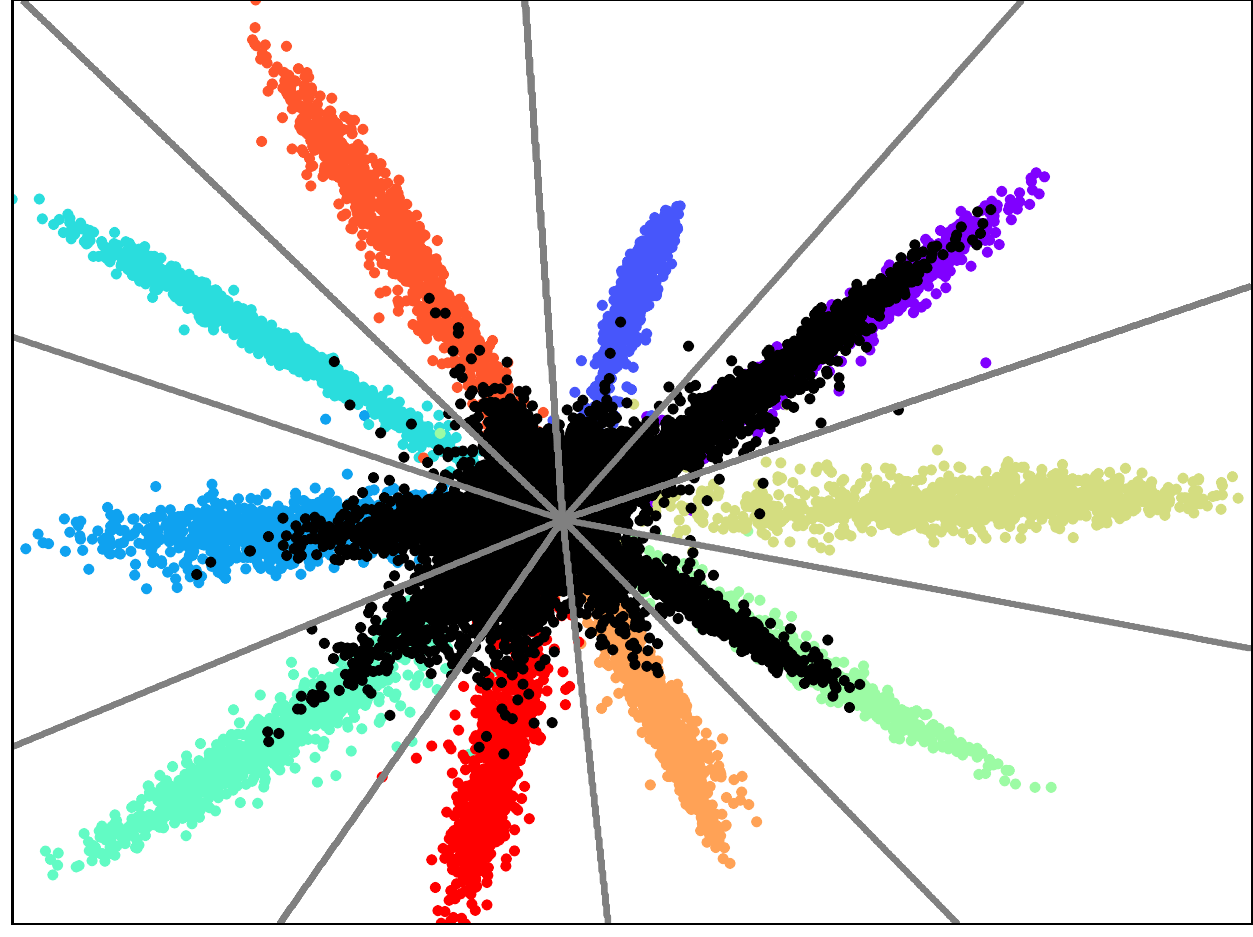}
        \caption{Objectosphere}
        \label{fig:lenet_objecto_omni}
    \end{subfigure}
    \begin{subfigure}[b]{0.25\textwidth}
        \centering
        \includegraphics[width=\textwidth]{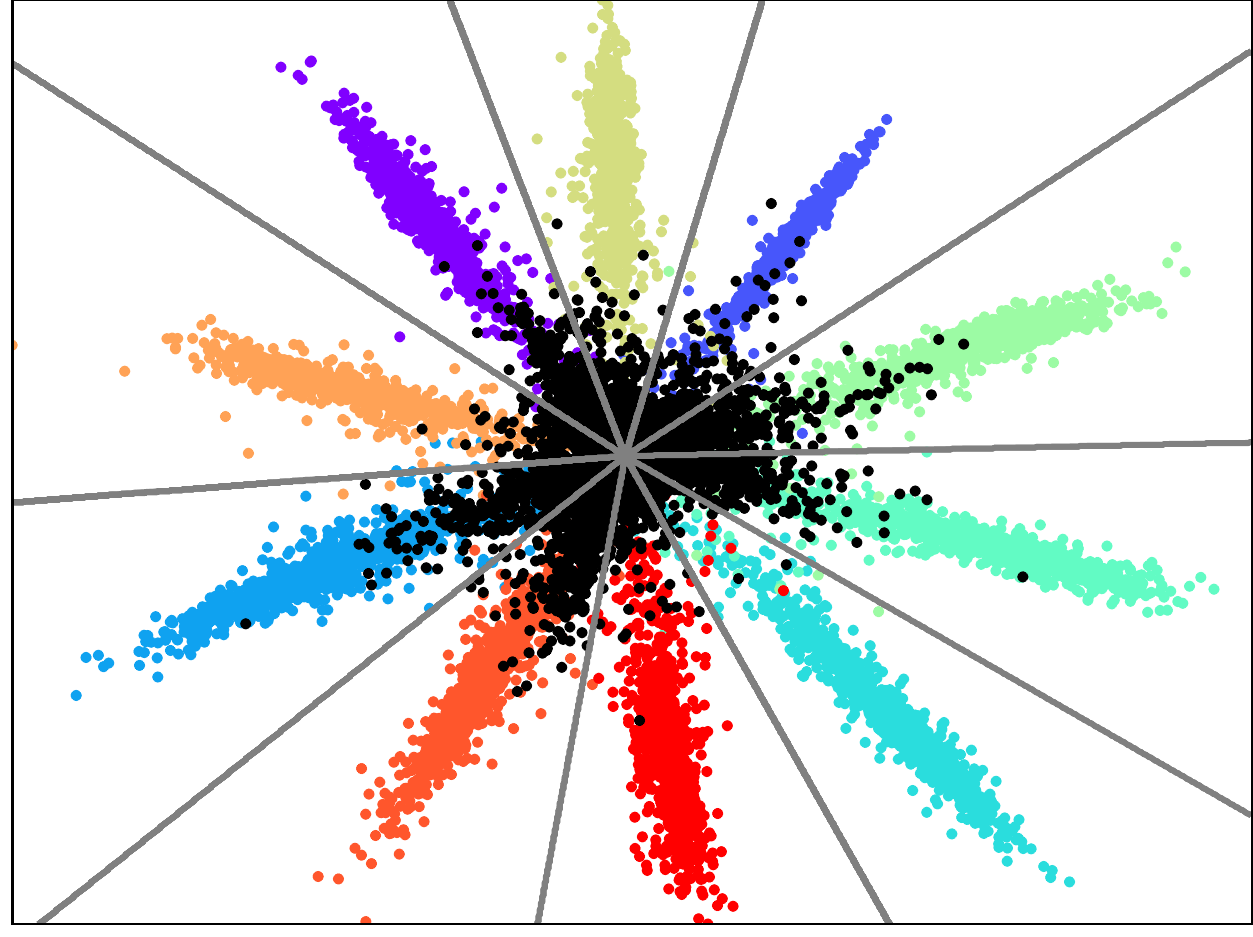}
        \caption{Tempered Mixup}
        \label{fig:lenet_tempered_mixup}
    \end{subfigure}
    \caption{2-D visualization of the effect of the different feature space regularization strategies on separating in-distribution and outlier inputs. The in-distribution training set is \textcolor{red}{M}\textcolor{orange}{N}\textcolor[HTML]{CCCC00}{I}\textcolor[HTML]{008000}{S}\textcolor{blue}{T}, while the unknown samples are from the \textbf{Extended-MNIST-Letters} dataset~\cite{cohen2017emnist}.  For Entropic Open Set and Objectosphere, the Omniglot dataset is used as a source for background samples.}
    \label{fig:lenet_comparison}
\end{figure}

As shown in Fig.~\ref{fig:lenet_comparison}, the Tempered Mixup model collapses the embedding of samples from the unknown classes towards the origin, thus reducing the overlap (and confusion with) known classes.  This is a dramatic improvement over common supervised training methods that improve model robustness including Label Smoothing~\cite{szegedy2016rethinking} and Center Loss~\cite{wen2016discriminative}. Tempered Mixup even improves on methods trained with an explicit background set such as Entropic Open Set and Objectosphere~\cite{dhamija2018reducing}.

\section{Experiments}
To evaluate the open set robustness of our method against current state-of-the-art techniques, we first compare against baselines established using the MNIST~\cite{lecun1998gradient} and CIFAR-10~\cite{krizhevsky2009learning} datasets as known classes and samples drawn from similar, but distinct, datasets as unknown classes. We then compare the performance of Tempered Mixup to standard Mixup training and other forms of VRM data augmentation.  Finally, we extend our small-scale experiments to show how our method scales to large-scale open set classification problems.

\subsection{Open Set Performance Assessment}
An open set classifier needs to correctly classify samples from known classes and identify samples from unknown classes. This makes evaluation more complex than out-of-distribution detection, which simplifies the detection task to a binary in/out classification problem.

Our primary metric is the area under the open set classification (AUOSC) curve. It measures the correct classification rate among known classes versus the false positive rate for accepting an open set sample, and has been used as a standard metric for open set classification~\cite{dhamija2018reducing}. The correct classification rate can be viewed as the difference between normal model accuracy and the false negative rate for rejection. Intuitively, AUOSC takes into account whether true positive samples are actually classified as the correct class and thus rewards methods which reject incorrectly classified positive samples before rejecting samples that are correctly classified. In addition to reporting the area under the curve, we also calculate the correct classification rate at a specific false positive rate of $10^{-1}$. Finally, we also report the area under the Receiver Operating Characteristic (AUROC) for identifying unknown classes as a measure of pure out-of-distribution detection performance. 

\subsection{Comparison Methods}

We compare our method against the following approaches:
\begin{enumerate}
    \item \textbf{Cross-entropy}: As a baseline, we train each network with standard cross-entropy loss to represent a common feature space for CNN-based models.
    
    \item \textbf{One-vs-Rest}~\cite{shu2017doc}: The one-vs-rest training strategy was implemented by substituting a sigmoid activation layer for the typical softmax activation and using a binary cross-entropy loss function.  In this paradigm, every image is a negative example for every category it is not assigned to. This creates a much larger number of negative training examples for each class than positive examples.
    
    \item  \textbf{Label Smoothing}~\cite{szegedy2016rethinking}: By smoothing target predictions during cross-entropy training, the model learns to regularize overconfident predictions and produce less confident and more calibrated predictions.
    
    \item \textbf{CenterLoss}~\cite{wen2016discriminative}: A form of model regularization that increases the robustness of the class-conditional feature representation by encouraging tightly grouped class clusters.  This is achieved by penalizing the Euclidean distance between samples and their class-mean (inter-class variance).  By reducing the inter-class variance, a more precise rejection threshold can theoretically be established for separating known from unknown samples.
    
    \item \textbf{Entropic Open Set}~\cite{dhamija2018reducing}: The Entropic Open Set method applies the confidence loss formulation (Eq.~\ref{eqn:conf_loss}) using a background class to train the model to reduce model confidence on unknown classes.
    
    \item \textbf{Objectosphere}~\cite{dhamija2018reducing}: Objectosphere separates known from unknown classes by training with a background class and reducing the magnitude of learned features for unknown classes.  Instead of using the confidence loss formulation, it uses a margin based hinge-loss centered around the origin in deep feature space. This loss reduces the magnitude of features from the background set and increases the magnitude of features for samples from known classes to be larger than a user defined margin. This method was shown to better separate known from unknown samples in network architectures with higher dimensional feature spaces.

\end{enumerate}
For all methods, identification of unknown classes is done by thresholding the maximum class posterior found by passing the model's output through a softmax activation. 

\subsection{Open Set Baselines}

We first study Tempered Mixup on common open set benchmarks. Following the protocol established in \cite{dhamija2018reducing}, the first baseline uses the LeNet++ CNN architecture~\cite{lecun1998gradient} with MNIST~\cite{lecun1998gradient} for known classes and a subset of the Extended-MNIST-Letters dataset~\cite{cohen2017emnist} for unknown classes.

The second benchmark uses a 32-layer Pre-Activation ResNet architecture~\cite{he2016deep} with CIFAR-10~\cite{krizhevsky2009learning} as known classes and 178 classes from TinyImageNet~\cite{le2015tiny} as unknown classes. TinyImageNet images were all resized to $32 \times 32$ images. We removed 22 classes from the original TinyImageNet dataset because they contained semantic overlap with CIFAR-10 classes based on hypernym or hyponyms, which were determined using the Wordnet lexical database~\cite{fellbaum2010wordnet}.

For both benchmarks, we train all open set classification methods on the known classes. Methods that use a background class (i.e., Entropic Open Set and Objectosphere) are additionally trained on the first 13 classes in the Extended-MNIST-Letters dataset.  For the CIFAR-10 baseline, a background class training set is drawn from non-overlapping classes in the CIFAR-100 dataset.  All methods are then evaluated on an even split of samples drawn from known and unknown classes. For MNIST, 10000 samples are used as a source of unknowns drawn from the final 13 classes in the Extended-MNIST-Letters dataset.  For CIFAR-10, 10000 samples are randomly selected from the TinyImageNet dataset as a source of unknowns.  

\begin{table}[t]
\centering
\caption{\textsc{Open-Set Classification Baselines}.  The correct classification rate at a false positive rate for open set classification of $10^{-1}$ and the areas under the resulting OSC and ROC curves.  In these experiments, the positive class used samples from the known set of classes seen during training and the negative class used samples from the unknown classes.  Best performance for each experiment and metric is in \textbf{bold}.}
\label{tab:small_scale_results}
\begin{tabular}{c|c|rrr}
\textit{Experiment} & \textit{Algorithm} & \multicolumn{1}{c}{\textit{\begin{tabular}[c]{@{}c@{}}CCR @ \\ FPR $10^{-1}$\end{tabular}}} & \multicolumn{1}{c}{\textit{AUOSC}} & \multicolumn{1}{c}{\textit{AUROC}} \\ \hline
\multirow{6}{*}{\begin{tabular}[c]{@{}c@{}} \underline{MNIST} \\ Unknown: \\ EMNIST-Letters \\ Arch: \\ LeNet++\end{tabular}} & Baseline & 0.7259 & 0.9066 & 0.9103 \\
 & One-vs-rest & 0.9556 & 0.9654 & 0.9814 \\
 & Label Smoothing & 0.8543 & 0.9315 & 0.9443 \\
 & CenterLoss & 0.9633 & 0.9695 & 0.9877 \\
 & Entropic Open-Set & 0.9712 & 0.9797 & 0.9892 \\
 & Objectosphere & 0.9570 & 0.9739 & 0.9801 \\
 & Tempered Mixup (Ours) & \textbf{0.9761} & \textbf{0.9821} & \textbf{0.9924} \\ \hline
\multirow{6}{*}{\begin{tabular}[c]{@{}c@{}} \underline{CIFAR-10}\\ Unknown: \\ TinyImageNet \\ Arch: \\ Pre-ResNet-32\end{tabular}} & Baseline & 0.5211 & 0.7694 & 0.8105 \\
 & One-vs-rest & 0.2363 & 0.7064 & 0.7559 \\
 & Label Smoothing & 0.0920 & 0.6841 & 0.7283 \\
 & CenterLoss & 0.4930 & 0.7613 & 0.8038 \\
 & Entropic Open-Set & 0.6766 & 0.7880 & 0.8344 \\
 & Objectosphere & 0.6720 & 0.8045 & \textbf{0.8584} \\
 & Tempered Mixup (Ours) & \textbf{0.6923} & \textbf{0.8099} & 0.8503
\end{tabular}%
\end{table}

\subsubsection{Results.}
Tempered Mixup achieves state-of-the-art-results in open set classification without the use of an explicit background class (see Table~\ref{tab:small_scale_results}). For the MNIST experiment, all methods show an improvement over the cross-entropy baseline. Tempered Mixup surpasses cross-entropy by 8\% in terms of AUOSC and even surpasses confidence loss methods that use a background set that is derived from the same dataset as the unknowns. For CIFAR-10, which uses a modern ResNet v2 architecture, Tempered Mixup achieves more than a 5\% improvement in terms of AUOSC over all methods that do not require a background set for training and is state-of-the-art in terms of AUOSC over all evaluated methods including those that train with an additional background set.

\subsection{Additional Evaluations}

\subsubsection{Performance Improvement from Target Rebalancing.}

\begin{figure}[t]
    \centering
    \begin{subfigure}{0.48\textwidth}
        \includegraphics[width=0.99\textwidth]{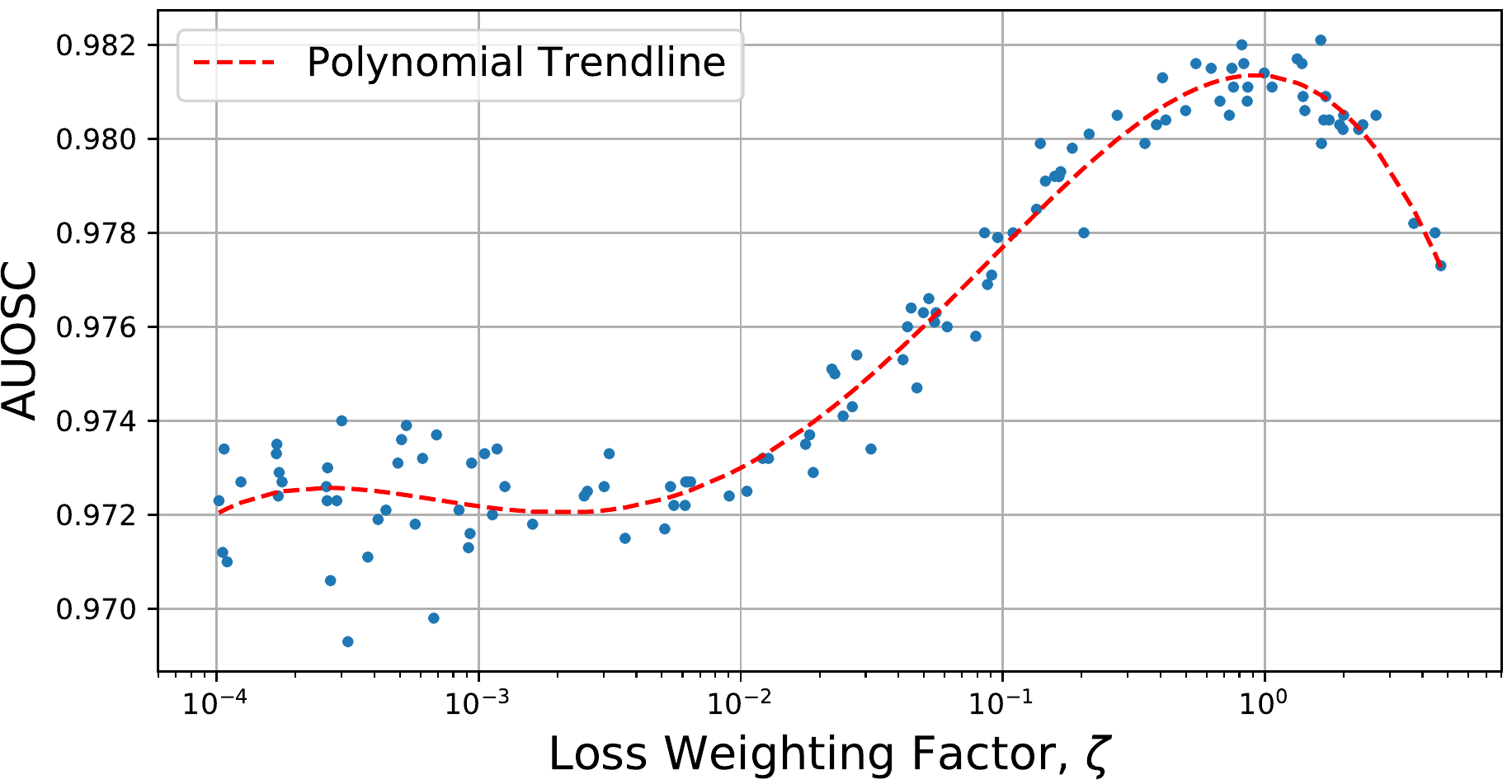}
        \caption{MNIST}
        \end{subfigure} \hfill
    \begin{subfigure}{0.48\textwidth}
        \includegraphics[width=0.99\textwidth]{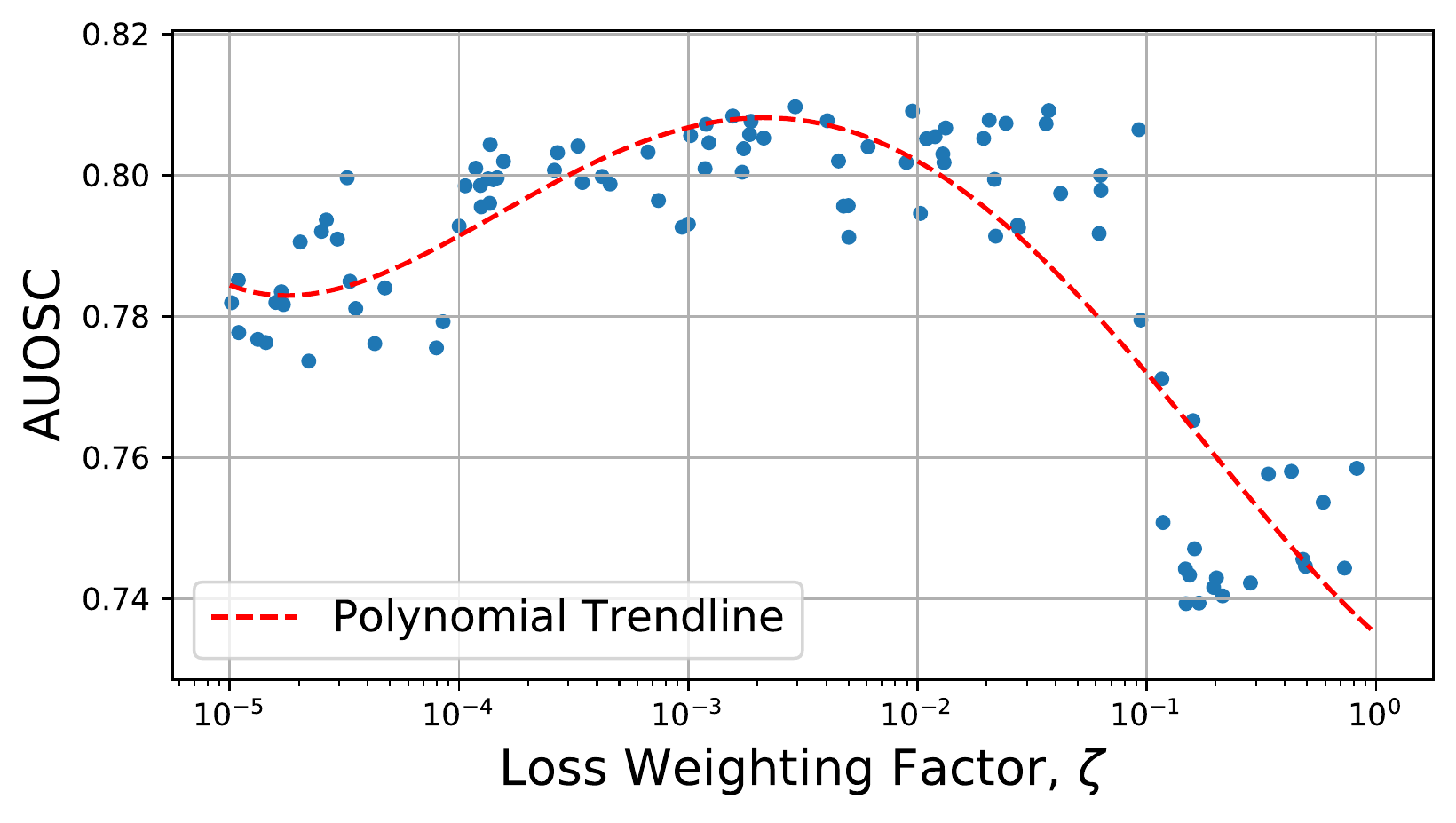}
        \caption{CIFAR-10}
        \end{subfigure}
    \caption{\textsc{Effect of Tempered Labels}: The effect of the tempering label smoothing loss term is varied to demonstrate the overall benefit on MNIST (left) and CIFAR-10 (right) Open Set Robustness.
    }
    \label{fig:mixup_coeff}
\end{figure}

 We seek to understand the benefit from target rebalancing towards improving the open set robustness gained from normal Mixup training.  To model this effect, we varied the weight applied to the label smoothing term using the loss weighting factor ($\zeta$ in Eq.~\ref{eq:tempered_Mixup_loss}). This allowed us to see the performance difference as the confidence loss portion of our formulation is emphasized over standard Mixup training (which is equivalent as $\zeta \rightarrow 0$).  As Fig.~\ref{fig:mixup_coeff} shows, performance improves as we increase the weight of the confidence loss until a point when the confidence loss prevents the model from achieving a high closed set accuracy, thus reducing the overall open set performance.

\subsubsection{Alternate Data Augmentation Schemes.} \label{sec:alt_formulations}
Mixup is part of a family of data augmentation approaches which work on the VRM principle.  As an alternative to the standard Empirical Risk Minimization formulation, VRM attempts to enlarge the support of the empirical training distribution by creating \textit{virtual} examples through various data augmentation schemes.  Other VRM data augmentation schemes have been proposed recently which have demonstrated increased robustness to certain forms of input corruption~\cite{chun2019empirical}; however, few have explicitly tested for open set robustness.

The Cutmix strategy~\cite{yun2019cutmix} overlays a random path from a separate training image and adjusts the target labels during training based on the ratio of the patch area to the original image.  Cutout~\cite{devries2017improved} is a similar variation, however the patch is made up of black (zero-valued) pixels.  We evaluated our Tempered Mixup formulation against these competing VRM schemes both in their normal formulation and with a tempered target label set where the entropy of the target distribution is adjusted based on the interpolation factor, $\lambda$.  As results show in Table~\ref{tab:mnist_cutmix_comp}, Tempered Mixup is superior to both Cutmix and Cutout and their tempered variants.

\begin{table}[t]
\centering
\caption{\textsc{Alternate VRM Comparison}. AUOSC performance of our method versus other VRM data augmentation techniques and their tempered variants.  Experiments use either MNIST or CIFAR-10 as known dataset and three different unknown datasets that vary in similarity to the known dataset. 
}
\label{tab:mnist_cutmix_comp}
\resizebox{\textwidth}{!}{%
\begin{tabular}{l|c|ccc}
\textit{Experiment} & \textit{Algorithm} & \textit{Gaussian Noise} & \textit{FMNIST} & \textit{EMNIST-Letters} \\ \hline
\multirow{5}{*}{\begin{tabular}[c]{@{}l@{}}\underline{MNIST}\end{tabular}} 
 & Baseline & \textbf{0.9878} & 0.9848 & 0.9066 \\
 & Cutmix & 0.8837 & 0.8751 & 0.8172 \\
 & Cutout & 0.9830 & 0.9813 & 0.9028 \\
 & Mixup & 0.9874 & \textbf{0.9875} & 0.9737 \\
 & Tempered Cutmix & 0.9805 & 0.9780 & 0.9249 \\
 & Tempered Cutout & 0.9844 & 0.9829 & 0.9121 \\
  & Tempered Mixup (Ours) & 0.9846 & \textbf{0.9875} & \textbf{0.9821} \\ \hline
\textit{Experiment} &  \textit{Algorithm} & \textit{SVHN} & \textit{LSUN} & \textit{Tiny ImageNet} \\ \hline
\multirow{5}{*}{\begin{tabular}[c]{@{}l@{}}\underline{CIFAR-10}\end{tabular}} 
 & Baseline & 0.8271 & 0.7934 & 0.7694 \\
 & Cutmix & 0.6249 & 0.7956 & 0.7697 \\
 & Cutout & 0.8174 & 0.7425 & 0.7414 \\
 & Mixup & 0.8193 & 0.7966 & 0.7886 \\
 & Tempered Cutmix & 0.7274 & 0.7803 & 0.7572 \\
 & Tempered Cutout & 0.7783 & 0.6963 & 0.7150 \\
 & Tempered Mixup (Ours) & \textbf{0.8340} & \textbf{0.8062} & \textbf{0.8099}
\end{tabular}%

}
\end{table}

\subsection{Large-Scale Open Set Classification}
Deployed systems typically operate on images with far higher resolution and many more categories than the open set baselines previously established for model regularization techniques. It is necessary to understand how well these systems work with higher resolution images and when the number of categories exceeds 100. As the number of categories increases, it can become increasingly difficult to identify a suitable set of background images for background regularization methods. 

To study open set classification for large-scale problems, we use the ImageNet Large-scale Visual Recognition Challenge 2012 dataset (ImageNet)~\cite{russakovsky2015imagenet}. ImageNet has 1.28 million training images (732-1300 per class) and 50000 labeled validation images (50 per class), which we use for evaluation.  All methods use an 18-layer ResNet CNN for classification~\cite{He_2016_CVPR} with an input image resolution of $224 \times 224$. In our experimental setup, the known set of classes consists of 500 classes from ImageNet. Following \cite{bendale2015towards}, unknown images for open set evaluation are drawn from categories of the 2010 ImageNet challenge that were not subsequently used and do not have semantic overlap with the 2012 ImageNet dataset. In total the open set dataset consisted of 16950 images drawn from the 339 categories.

We compare Tempered Mixup against the baseline cross-entropy method, Objectosphere, and a combination that incorporates both Objectosphere and Tempered Mixup training. Objectosphere was chosen because it is the best method for using a background dataset based on our previous experiments. To our knowledge, Objectosphere has not been previously evaluated on large-scale problems, thus for a background training set, we use 1300 images from the Places scene understanding validation dataset~\cite{zhou2017places}. We again ensure that all classes do not have semantic overlap with any ImageNet category in either the known or unknown evaluation set, as verified by hypernym and hyponym relationship lookup in the Wordnet lexical database. All models are trained using SGD with a mini-batch size of 256, momentum weighting of 0.9, and weight decay penalty factor of 0.0001 for 90 epochs, starting with a learning rate of 0.1 that is decayed by a factor of 10 every 30 epochs. The baseline cross-entropy trained model for the 500 class partition achieves 78.04\% top-1 (94.10\% top-5) accuracy.

\subsubsection{Results.}
\begin{figure}[t]
    \centering
    \includegraphics[width=\textwidth]{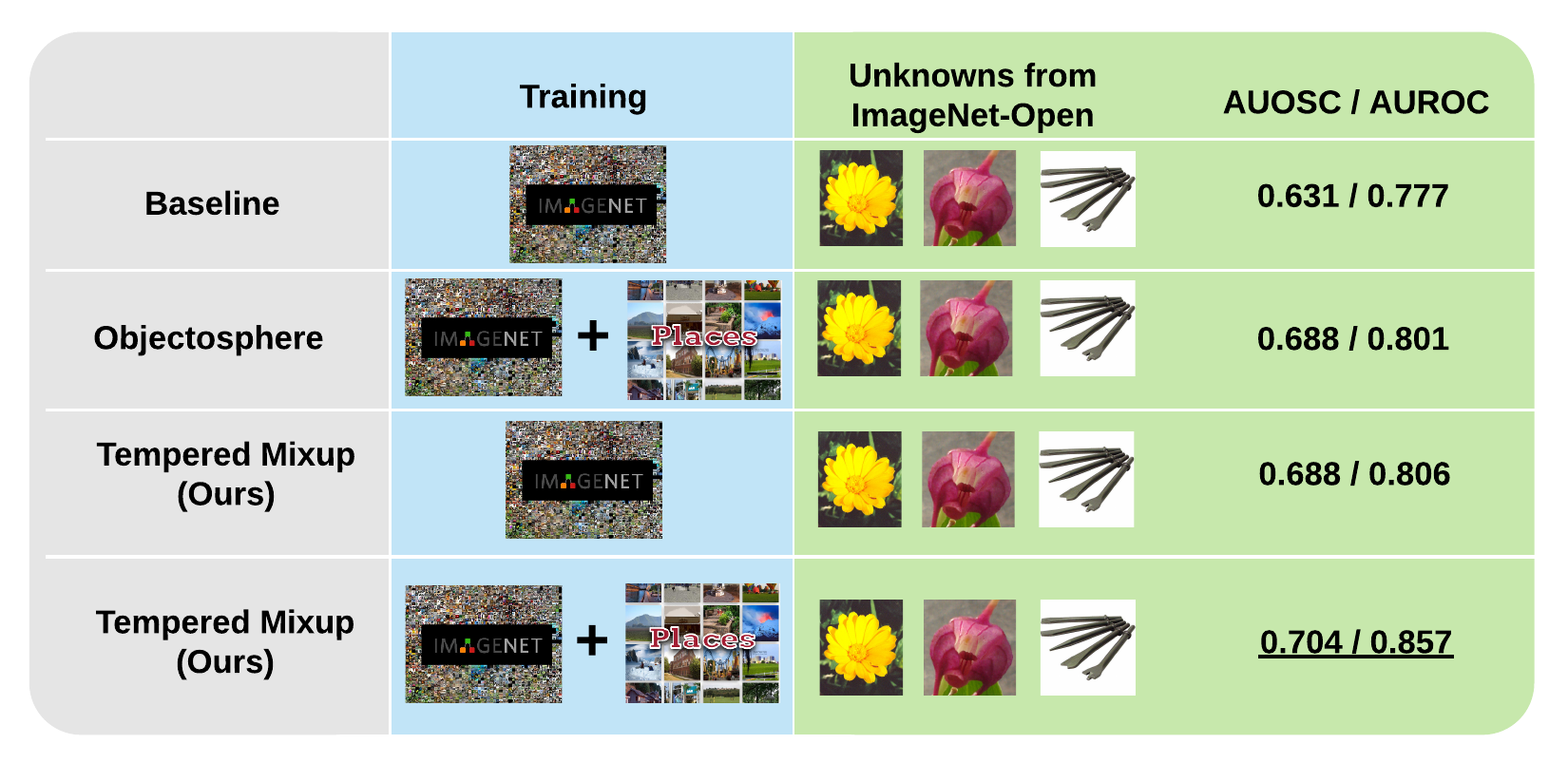}
    \caption{\textsc{Large-Scale Open Set Classification}. Training data is made up of either ImageNet only or ImageNet plus background images from the Places dataset.  Unknowns for evaluation are drawn from the ImageNet-Open dataset.}
    \label{fig:large_scale}
\end{figure}

The results from our ImageNet experiments are shown in Fig.~\ref{fig:large_scale}. We compute the AUOSC metric using the top-1 correct classification rate and report AUROC as a measure of OOD detection capability.  For this large-scale experiment, Tempered Mixup shows roughly the same open set robustness as compared to Objectosphere without having to train with an additional dataset of background samples.  To try and gain even better open set performance we augmented our Tempered Mixup formulation with the same background samples used in the Objectosphere training and a uniform distribution target among the known classes for these samples.  In this way, our model trains on multiple combinations of mixed up samples, including combinations of known and unknown classes. The resulting hybrid model achieved the best open set performance over either the Tempered Mixup or the Objectosphere methods alone.

\section{Discussion}
The results from both small-scale and large-scale open set classification problems are evidence that Tempered Mixup is an effective means of improving open set robustness through feature space regularization without having to train with a source of representative unknown samples.  We additionally have shown that when a representative background class is available, samples can easily be added into the training pipeline to gain additional robustness.

Our ImageNet open set classification results show that our Tempered Mixup formulation is as effective alone as training with a background class  without the additional overhead.  When an effective background class is available, we have also demonstrated that our formulation can take advantage of this additional training data to further improve open set robustness.  In this case when the nature of the unknown classes to be rejected is known to a degree that an effective background class can be procured, then this is equivalent to hard negative mining for training with a confidence loss framework to reduce the network activation towards these unknown samples.

While we have demonstrated that our training paradigm builds a robust feature representation that improves model robustness in detecting novel classes unseen during training, this property is only tested in our work with a baseline confidence thresholding inference method.  More advanced inference methods could easily be applied to our models to yield even better open set performance. 


\section{Conclusion}
In this paper we developed a novel technique for improving the feature space of a deep CNN classifier to enable better robustness towards samples from unknown classes.  We combined the concept of Mixup augmentation with a novel formulation of confidence loss to train a CNN to produce less confident predictions for samples off of the input distribution defined by the training set.  Experimental evidence shows that this formulation performs favorably against current state-of-the-art methods including confidence loss regularization with a background class. This strategy could be especially useful when an appropriate background set is not available in large-scale, real-world classification environments.  

\ifthenelse{\boolean{acknowledgement}}{
\paragraph{Acknowledgements.}
This work was supported in part by NSF award \#1909696, the DARPA/MTO Lifelong Learning Machines program [W911NF-18-2-0263], and AFOSR grant [FA9550-18-1-0121]. The views and conclusions contained herein are those of the authors and should not be interpreted as representing the official policies or endorsements of any sponsor.
}

\end{document}